\documentclass[fleqn,10pt]{wlscirep}
\usepackage[utf8]{inputenc}
\usepackage[T1]{fontenc}

\usepackage{algorithmic}
\usepackage{algorithm}
\usepackage[caption=false,font=normalsize,labelfont=sf,textfont=sf]{subfig}

\usepackage{multirow}
\usepackage{array}
\newcolumntype{M}[1]{>{\centering\arraybackslash}m{#1}}

\usepackage{caption}
\usepackage{float}

\title{GSVD-NMF: Recovering Missing Features in Non-negative Matrix Factorization }

\author[1]{Youdong Guo}
\author[1,*]{Timothy E. Holy}

\affil[1]{Washington University in St. Louis, Department of Neuroscience, St. Louis, 63130, USA}

\affil[*]{holy@wustl.edu}


\begin{abstract}
Non-negative matrix factorization (NMF) is an important tool in signal processing and widely used to separate mixed sources into their components. 
Algorithms for NMF require that the user choose the number of components in advance, and if the results are unsatisfying one typically needs to start again with a different number of components. To make NMF more interactive and incremental, here we introduce GSVD-NMF, a method that proposes new components based on the generalized singular value decomposition (GSVD) to address discrepancies between the initial under-complete NMF results and the SVD of the original matrix.
Simulation and experimental results demonstrate that GSVD-NMF often effectively recovers multiple missing components in under-complete NMF, with the recovered NMF solutions frequently reaching better local optima. The results further show that GSVD-NMF is compatible with various NMF algorithms and that directly augmenting components is more efficient than rerunning NMF from scratch with additional components.
By deliberately starting from under-complete NMF, GSVD-NMF has the potential to be a recommended approach for a range of general NMF applications.
\end{abstract}

\begin{document}

\flushbottom
\maketitle
%
%
\thispagestyle{empty}


\section*{Introduction}

Non-negative matrix factorization(NMF) is an effective tool for learning latent features from data in circumstances where plausible features are constrained to be nonnegative. 
It has been widely used in the analysis of many types of data \cite{lee1999learning, rajabi2014spectral, wang2016hypergraph, chen2019experimental, kowsari2019text, gemmeke2013exemplar, leplat2020blind, dubroca2012weighted, lin2020optimization, mcclure2014novel}.
The NMF is given by
\begin{align}
    \begin{aligned}
        \label{NMF_eq}
        \mathbf{X} \approx \mathbf{\mathbf{WH}}\  \textrm{subject to}\ \ \mathbf{\mathbf{W}}\geq 0,\ \mathbf{H}\geq0
    \end{aligned}
\end{align}
where $\mathbf{\mathbf{W}}\in\mathbb{R}^{m\times r_0}$ and $\mathbf{H}\in\mathbb{R}^{r_0\times n}$ are two nonnegative matrices. 
$r_0$ is the number of components, which ideally represents the actual number of features in the data matrix. 
To obtain $\mathbf{W}$ and $\mathbf{H}$ in equation~(\ref{NMF_eq}), here we focus on minimizing the square-difference of the original matrix ($\mathbf{X}$) and the factorization ($\mathbf{WH}$)
 \begin{align}
    \begin{aligned}
    \label{SED_obj}
     D_E(\mathbf{X}||\mathbf{W}\mathbf{H})=\frac{1}{2}\left\lVert \mathbf{X}-\mathbf{WH}\right\rVert^2_F
    \end{aligned}
\end{align}
This Squared Euclidean Distance is one of the most widely used objective functions for NMF, and a variety of algorithms have been developed to minimize it \cite{cichocki2009fast, lin2007projected, lee1999learning, gillis2010using}.
Although NMF is widely used, in some circumstances it can fail to identify an accurate and comprehensive representation of the data features. 
Such failures arise from three sources, including the choice of $r_0$, NP-hardness, and non-uniqueness \cite{gillis2014and}.
Here we develop methods to incrementally change $r_0$ and then show that these techniques represent a powerful strategy for improving the quality of solutions.

We say that NMF is ``under-complete,'' with too few components in $\mathbf{W}$ and $\mathbf{H}$, if $r_0$ is smaller than the number features in $\mathbf{X}$.
In such cases, features may be missed or integrated into other components.
We recognize that in real-world signal processing contexts, the correct number of components may be undefined, and the usefulness of the decomposition is assessed in part by how well the resulting components explain the underlying phenomena under study.
Various methods have been proposed to learn $r_0$, but no known procedure results in an unambiguous answer\cite{lin2020optimization, gillis2014and, fogel2023rank}. 
Because performing NMF from beginning with a different number of components can be time-consuming, ideally one would have an efficient method to manipulate/augment the components in a pre-existing under-complete NMF solution. 
Among the methods mentioned above for minimizing equation~(\ref{SED_obj}), rank-one underapproximation \cite{gillis2010using, biggs2008nonnegative} incrementally adds one column (component) in $\mathbf{W}$ and $\mathbf{H}$, and then re-optimizes the augmented factorization.
However, these methods can only add one component at a time, necessitating iterative augmentation to introduce multiple components, which reduces reconstruction efficiency, particularly for large-scale datasets. 
Moreover, these methods update new components based on the residual between the original matrix and the obtained components. 
This residual is required to be non-negative, yet the residual between the original matrix and a typical under-complete NMF solution is not guaranteed to be non-negative. 
Additionally, due to these constraints, the methods become computationally expensive, and their convergence to a stationary point remains undemonstrated \cite{kim2014algorithms}.
Sparse regression-based NMF pruning begins with an excessive number of components and removes unnecessary ones through sparsity constraints\cite{hoyer2004non}. 
These methods do not heavily rely on robust estimation of the number of components, as unnecessary components are effectively reduced to zero. 
However, they introduce an additional sparsity coefficient, which must be learned or tuned separately. 
Moreover, different approaches may produce varying results when applied to real-world datasets\cite{morup2009tuning,hinrich2018probabilistic,li2013non}.

For the second issue mentioned above, since NMF is not convex for $\mathbf{W}$ and $\mathbf{H}$ simultaneously, NMF may converge to local minimum or stop at saddle points. 
In this case, NMF would fail to give a good representation of features in $\mathbf{X}$ even if the ``correct'' $r_0$ exists and is known.
Algorithms have been developed to mitigate this issue by either focusing on special types of matrices for which the problem is convex\cite{arora2012computing, mcclure2014novel}, by improving initialization to increase the likelihood of landing in high-quality minima\cite{boutsidis2008svd, esposito2021review, fathi2023initialization, casalino2014subtractive, alshabrawy2012underdetermined, atif2019improved}, and/or by adding regularization terms to the objective that steer optimization in desirable directions\cite{yuan2020convex, huang2020robust, ince2022weighted, jiao2020hyper, deng2022graph}.
However, no algorithm can guarantee the global optimum in the general case.

The Generalized Singular Value Decomposition (GSVD) is an extension of the standard SVD, used for simultaneously decomposing two matrices that may have different dimensions but share a common column space. 
It has applications in areas such as signal processing, multivariate statistics, and data analysis, particularly when comparing two datasets or solving generalized eigenvalue problems.
It is especially useful when analyzing the relationship between two datasets, as it finds conjugate directions that maximize the ratio of corresponding components from two different matrices\cite{callaerts1990comparison}.

In this paper, we present a method called GSVD-NMF to augment components for an existing under-complete NMF solution.
Intuitively, given that SVD provides a globally optimal matrix factorization, and GSVD identifies conjugate directions that maximize the signal-to-signal ratio, leveraging the GSVD between the existing NMF results and the SVD of $\mathbf{X}$ may recover missing components for under-complete NMF.
Directions with generalized singular value different from one correspond to discrepancies between the NMF and the SVD, and thus may present opportunities to improve the existing NMF. 
We propose a deterministic strategy that leverages GSVD results to augment the existing factorization. 
We employ NNDSVD truncation\cite{boutsidis2008svd} to maintain the non-negativity of the augmented components.
Using nonnegative least squares, we also modulate the amplitude of ``old'' components, with the aim of providing more opportunity for the new components to contribute substantively to the factorization.
The entire pipeline is tested through simulations and on multiple real-world datasets with different NMF algorithms for source separation. 
The results demonstrate that our method effectively initializes new components for under-complete NMF, with the augmented solutions achieving better local optima and greater efficiency than standard NMF approaches.
Moreover, starting from under-complete NMF (deliberately setting rank number $r_0$ smaller than the number of features in $\mathbf{X}$), the proposed method can help NMF converge to better local optima. 
Finally, the proposed method allows one to efficiently expand the number of components, which can be convenient and effective for interactive analysis of large-scale data.
It should be noted that this paper only discusses augmenting new components for under-complete NMF. When $r_0$ exceeds the desired number of components, resulting in over-complete NMF, the components need to be recombined to optimize NMF. This topic is covered in a separate paper\cite{guo2024optimal}.
\section*{Methods}
\label{method}
\begin{figure*}[!t]
    \centering
    \includegraphics[width=6.93in]{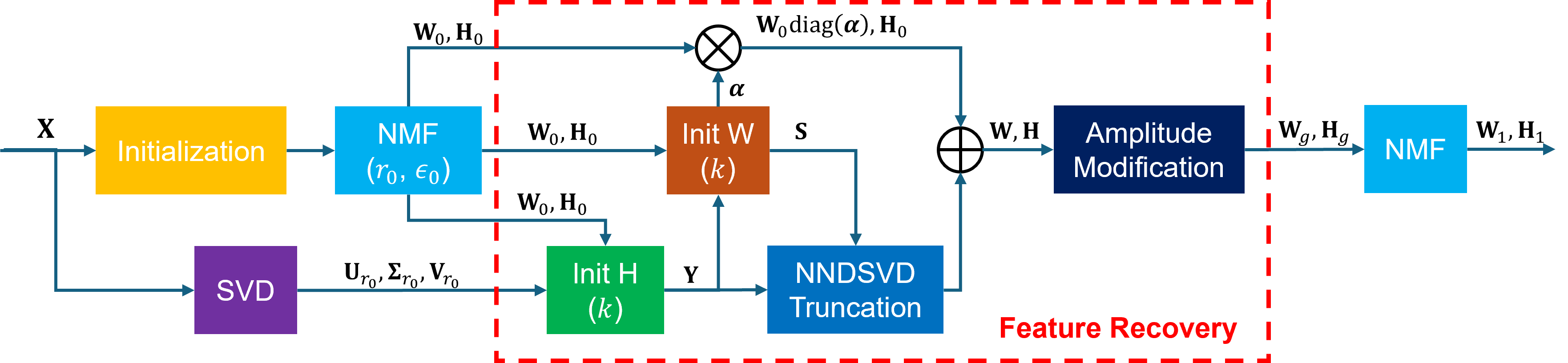}
    \caption{The whole pipeline of GSVD-NMF.}
    \label{whole_pipe}
\end{figure*}

    The whole pipeline of the proposed method is given in Fig. \ref{whole_pipe}. 
    Given that standard NMF typically yields local optima and SVD constitutes a global optimum of matrix factorization, our core concept involves identifying features captured by SVD but absent in NMF outcomes. 
    These identified features are then incorporated as new components into the existing NMF results. $\mathbf{W}_0$ and $\mathbf{H}_0$ are the results of under-complete NMF with $r_0$ components. 
    We use $\mathbf{U}_{r_0}$, $\boldsymbol{\Sigma}_{r_0}$ and $\mathbf{V}_{r_0}$ to denote the results of rank-$r_0$ approximate SVD, which is the global optimum of rank-$r_0$ matrix factorization of $\mathbf{X}$. 
    Using the GSVD, we compare the results of under-complete NMF and rank-$r_0$ SVD, and on that basis propose new rows (denoted $\mathbf{Y}$) for $\mathbf{H}$ and new columns (denoted $\mathbf{S}$) for $\mathbf{W}$.
    The matrices $\mathbf{S}$ and $\mathbf{Y}$ are not guaranteed to be non-negative, so NNDSVD truncation\cite{boutsidis2008svd} is applied to preserve the non-negative approximation.
    Additionally, the amplitudes of the original and newly added components are jointly adjusted using a non-negative least-squares approach to balance the power between the existing and new components.
    Finally, all components are polished by NMF to produce the final factorization. We now describe these steps in detail.

\begin{figure*}[!t]
    \centering
    \includegraphics[width=6.9in]{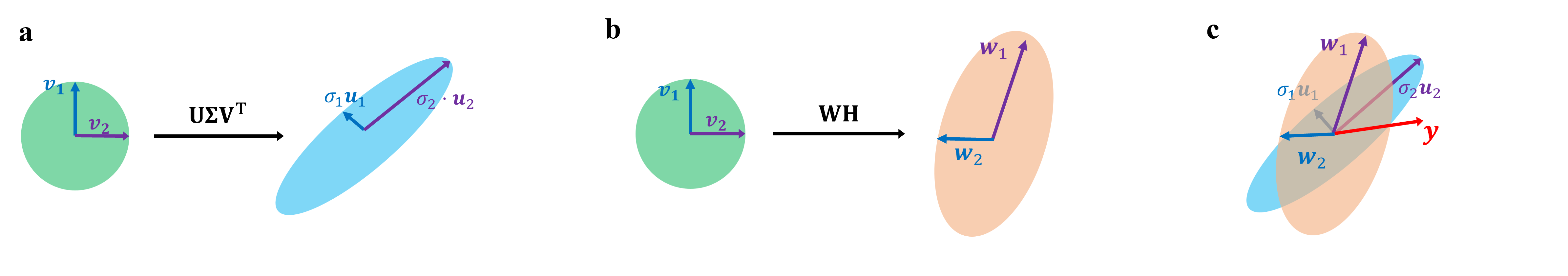}
    \caption{Concept of GSVD-based feature recovery for NMF (2$\times$2 case). (a) Under multiplication by $\mathbf{X}$, points in the green disk map to points in the blue ellipse; the key elements of $\mathbf{X}$'s SVD are denoted. (b) An inexact NMF factorization maps to a different ellipse (tan). (c) GSVD-NMF suggests new directions ($\mathbf{y}$) to make the tan ellipse more like the blue ellipse.}
    \label{fig_gsvd}
\end{figure*}

    Our concept is most easily understood by viewing matrices as linear transformations that act on vectors in the domain space, which in Fig. \ref{fig_gsvd} will be represented as points in the unit disc (green). Fig. \ref{fig_gsvd}(a) represents the action of the SVD of $\mathbf{X}$ on a unit disc and Fig. \ref{fig_gsvd}(b) is action of the NMF of $\mathbf{X}$ on the unit disc, projected into the subspace spanned by the SVD.
    Any difference between the resulting ellipses in the range-space (Fig. \ref{fig_gsvd}(c)) represents a mismatch between the NMF and the optimal rank-$r_0$ factorization.
    Our primary objective is to identify new directions that can reshape the NMF solution to more closely approximate the SVD solution, as exemplified by the vector $\mathbf{y}$ in Fig. \ref{fig_gsvd}(c). 
 
 As mentioned above, the GSVD of a pair of matrices identifies conjugate directions that maximize the oriented signal-to-signal ratio. Naively, similar to maximizing variance for principal component analysis (PCA), one might imagine that the new direction $\mathbf{y}\in\mathbb{R}^{n \times1}$ could be obtained by optimizing
\begin{align}
    \begin{aligned}
    \label{max_obj}
    \max_{\mathbf{y}}\frac{\lVert \mathbf{U}_{r_0} \boldsymbol{\Sigma}_{r_0} \mathbf{V}_{r_0}^\mathrm{T}\mathbf{y}\rVert^2}{\lVert \mathbf{W}_0\mathbf{H}_0\mathbf{y}\rVert^2}
    \end{aligned}
\end{align}
However, this suffers from a conceptual problem: equation~(\ref{max_obj}) is infinite for any $\mathbf{y}$ in the null space of $\mathbf{W}_0\mathbf{H}_0$ but not in the null space of the SVD. 
As there are at least $\mathrm{min}(m,n)-r_0$ dimensions in the nullspace of $\mathbf{W}_0\mathbf{H}_0$, any discrepancy between the SVD and NMF domain subspaces provides an infinite number of potential solutions for equation~(\ref{max_obj}).
Thus, we model $\mathbf{y}$ in the column space of $\mathbf{V}$ and instead optimize a variant of equation~(\ref{max_obj}) that projects the NMF result down to the SVD subspace:
\begin{align}
    \begin{aligned}
    \label{max_obj_2}
    \max_{\hat{\mathbf{y}}}\frac{\lVert \mathbf{U}_{r_0}\boldsymbol{\Sigma}_{r_0} \hat{\mathbf{y}}\rVert^2}{\lVert \mathbf{W}_0\mathbf{H}_0\mathbf{V}_{r_0}\hat{\mathbf{y}}\rVert^2}
    \end{aligned}
\end{align}
where $\hat{\mathbf{y}}\in\mathbb{R}^{r_0 \times1}$ and $\mathbf{y}=\mathbf{V}_{r_0}\hat{\mathbf{y}}$.
This is equivalent to optimizing the generalized Rayleigh Quotient. 
 Since $\boldsymbol{\Sigma}_{r_0}^\mathrm{T}\boldsymbol{\Sigma}_{r_0}$ and $\mathbf{V}_{r_0}^\mathrm{T}\mathbf{H}_0^\mathrm{T}\mathbf{W}_0^\mathrm{T}\mathbf{W}_0\mathbf{H}_0\mathbf{V}_{r_0}$ are symmetric, the extreme values of equation~(\ref{max_obj_2}) satisfy a generalized eigenvalue problem\cite{li2015rayleigh}
\begin{align}
    \begin{aligned}
        \label{gen_eigen_final}
        \boldsymbol{\Sigma}_{r_0}^\mathrm{T}\boldsymbol{\Sigma}_{r_0}\hat{\mathbf{y}}
        =\lambda \mathbf{V}_{r_0}^\mathrm{T}\mathbf{B}^\mathrm{T}\mathbf{B}\mathbf{V}_{r_0}\hat{\mathbf{y}}
    \end{aligned}
\end{align}
where $\mathbf{B} = \mathbf{W}_0\mathbf{H}_0$. Equation~(\ref{gen_eigen_final}) can be solved to higher numerical precision by the GSVD of $\boldsymbol{\Sigma}_{r_0}$ and $\mathbf{B}\mathbf{V}_{r_0}$, resulting in
    \begin{align}
        \begin{aligned}
        \label{GSVD_solution}
            &\boldsymbol{\Sigma}_{r_0} = \mathbf{M}_1 \mathbf{D}_1 \mathbf{Q}^\mathrm{T}\\
            &\mathbf{B}\mathbf{V}_{r_0} = \mathbf{M}_2\mathbf{D}_2\mathbf{Q}^\mathrm{T}
        \end{aligned}
    \end{align}
    The matrices $\mathbf{M}_1$, $\mathbf{M}_2 \in \mathbb{R}^{r_0\times r_0}$ are unitary and $\mathbf{Q}\in \mathbb{R}^{r_0\times r_0}$. Here, the matrices $\mathbf{D}_1$, $\mathbf{D}_2 \in \mathbb{R}^{r_0\times r_0}$ are given by
        \begin{align}
        \begin{aligned}
        \label{D_1andD_2}
                \mathbf{D}_1  = 
                \begin{pmatrix}
                    \mathbf{I} & \mathbf{0}\\
                    \mathbf{0} & \mathbf{C}
                \end{pmatrix},\ \ 
               \mathbf{D}_2  = 
                \begin{pmatrix}
                    \mathbf{0} & \mathbf{G}\\
                    \mathbf{0} & \mathbf{0}
                \end{pmatrix} 
        \end{aligned}
    \end{align}
where $\mathbf{C}, \mathbf{G}\in \mathbb{R}^{l\times l}$ are real, diagonal matrices and $l$ denotes the rank of $\mathbf{B}\mathbf{V}_{r_0}$. $\mathbf{I} \in \mathbb{R}^{(r_0-l) \times (r_0-l)}$ is identity matrix.  
If $l=r_0$, $\mathbf{D}_1=\mathbf{C}$ and $\mathbf{D}_2=\mathbf{G}$.
    Plugging equation~(\ref{GSVD_solution}) into  equation~(\ref{gen_eigen_final}) yields
    \begin{align}
        \begin{aligned}
        \mathbf{D}_1^\mathrm{T}\mathbf{D}_1\mathbf{Q}^\mathrm{T}\hat{\mathbf{y}}=\lambda \mathbf{D}_2^\mathrm{T}\mathbf{D}_2\mathbf{Q}^\mathrm{T}\hat{\mathbf{y}}
        \end{aligned}
    \end{align}
    Letting $\mathbf{z} = \mathbf{Q}^\mathrm{T}\hat{\mathbf{y}}$, we need to solve 
     \begin{align}
        \begin{aligned}
        \label{final_normal_eigen}
        \mathbf{D}_1^\mathrm{T}\mathbf{D}_1\mathbf{z}=\lambda\mathbf{D}_2^\mathrm{T}\mathbf{D}_2\mathbf{z}
        \end{aligned}
    \end{align}
    $\mathbf{D}_1^\mathrm{T}\mathbf{D}_1$ and $\mathbf{D}_2^\mathrm{T}\mathbf{D}_2$ are real, non-negative diagonal matrices. 
    Assume that $\mathbf{D}_1^\mathrm{T}\mathbf{D}_1 = \mathrm{diag}(d_{11}^2, \ldots, d_{1r_0}^2)$ and  $\mathbf{D}_2^\mathrm{T}\mathbf{D}_2 = \mathrm{diag}(d_{21}^2, \ldots, d_{2r_0}^2)$. 
    If the rank of $\mathbf{B}$ is $r_0$ ($l=r_0$), all generalized singular values $d_{1i}/d_{2i}$ are finite; conversely, if $l < r_0$, the generalized singular values are infinite for $i = 1, \ldots, r_0-l$ and finite thereafter. When $d_{1i}/d_{2i}$ is finite ($d_{2i}\neq0$), the corresponding $\mathbf{z}_i$ are the unit coordinate vectors. 
     Thus, $\hat{\mathbf{y}}_i = \left(\mathbf{Q}^\mathrm{T}\right)^{-1}\mathbf{z}_i$ and the new directions are given by
    \begin{align}
        \begin{aligned}
        \label{y_i}
            \mathbf{y}_{i} = \mathbf{V}_{r_0}\left(\mathbf{Q}^\mathrm{T}\right)^{-1}\mathbf{z}_i
        \end{aligned}
    \end{align}
    $\mathbf{y}_{i}$ is the $i$-th column of $\mathbf{V}_{r_0}\left(\mathbf{Q}^\mathrm{T}\right)^{-1}\ \mathrm{for}\ i = r_0-l+1,\dots, r_0$. When $i = 1,\dots, r_0-l$, $d_{1i}/d_{2i}$ is infinite ($d_{2i}=0$), we similarly select $\mathbf{y}_{i}$ using the $i$-th column of $\mathbf{V}_{r_0}\left(\mathbf{Q}^\mathrm{T}\right)^{-1}$. 
    These represent directions that are missing entirely from the $\mathbf{W}_0\mathbf{H}_0$ factorization. 
    
    Typically, GSVD-NMF has the capacity to suggest missing directions only for rank-2 NMF and higher. For rank-1 factorizations, NMF algorithms find the global optimum, and since the first component in the SVD of a non-negative matrix is non-negative, the rank-1 NMF solution should already be equivalent to the first SVD component. 

    After fixing $\mathbf{Y}$, the corresponding additional components for $\mathbf{W}_0$ (defined as $\mathbf{S}$) are derived from a least-squares problem
     \begin{align}
        \begin{aligned}
        \label{init_W}
            \min_{\mathbf{S}, \boldsymbol{\alpha}}\lVert \mathbf{X}-\sum_{p=1}^{r_0}\alpha_p\mathbf{w}_{0p}\mathbf{h}_{0p}^\mathrm{T}-\mathbf{SY}\rVert^2\\
        \end{aligned}
    \end{align}
    where $\mathbf{w}_{0p}$ is the $p$-th column in $\mathbf{W}_0$, $\mathbf{h}_{0p}$ is the $p$-th row in $\mathbf{H}_0$ and $\boldsymbol{\alpha} = [\alpha_1, \alpha_2, \dots, \alpha_{r_0}]$. 
    Here $\boldsymbol{\alpha}$ is required to be non-negative in order to preserve the non-negativity of the original
components $\mathbf{W}_0\mathbf{H}_0$.
    The vector $\boldsymbol{\alpha}$ is introduced to modulate the magnitude of the pre-existing components, thereby improving the adaptability of the matrix $\mathbf{S}$ in the minimization of equation~(\ref{init_W}). Equation~(\ref{init_W}) can be expanded to the quadratic form
    \begin{align}
        \begin{aligned}
        \label{convex_init_W}
            E =& \lVert \mathbf{X}-\sum_{p=1}^{r_0}\alpha_p\mathbf{w}_{0p}\mathbf{h}_{0p}^\mathrm{T}-\mathbf{SY}\rVert^2\\
              =& \boldsymbol{\alpha}^\mathrm{T}\boldsymbol{\Theta}\boldsymbol{\alpha}-2\boldsymbol{\xi}^\mathrm{T}\boldsymbol{\alpha}+\Phi
                -2\boldsymbol{\gamma}^\mathrm{T}\mathbf{m}
                +2\boldsymbol{\alpha}^\mathrm{T}\mathbf{P}\mathbf{m}+\mathbf{m}^\mathrm{T}\boldsymbol{\Psi}\mathbf{m}
        \end{aligned}
    \end{align}
    where $\mathbf{m}=\left[\mathbf{s}_1^\mathrm{T}, \mathbf{s}_2^\mathrm{T}, \dots, \mathbf{s}_k^\mathrm{T}\right]^\mathrm{T}$ represents $\mathbf{S}$ as a long vector, $\boldsymbol{\Theta}\in \mathbb{R}^{r_0\times r_0}$ with  
    $\Theta = \left(\mathbf{W}_0^\mathrm{T}\mathbf{W}_0\right)\odot\left(\mathbf{H}_0\mathbf{H}_0^\mathrm{T}\right)$.
    $\boldsymbol{\xi}\in\mathbb{R}^{r_0\times 1}$ with $\xi_p=\mathbf{w}_{0p}^\mathrm{T}\mathbf{X}\mathbf{h}_{0p}$, $\Phi=\sum_{i,j}X_{ij}$, $\boldsymbol{\gamma}=\left[(\mathbf{X}\mathbf{y}_1)^\mathrm{T}, (\mathbf{X}\mathbf{y}_2)^\mathrm{T}, \dots, (\mathbf{X}\mathbf{y}_k)^\mathrm{T}\right]^\mathrm{T}$. $\mathbf{P}$ and $\boldsymbol{\Psi}$ are a block matrices, where $\mathbf{P}_{pp'}=\mathbf{h}_{0p}^\mathrm{T}\mathbf{y}_{p'}\mathbf{w}_{0p}^\mathrm{T}$ and $\boldsymbol{\Psi}_{pp'}=\mathbf{y}_p^\mathrm{T}\mathbf{y}_{p'}\mathbf{I}_m$, $\mathbf{I}_m\in\mathbb{R}^{m\times m}$ is an identity matrix. The stationary point of $E$ with respect to $\mathbf{m}$ is 
    \begin{align}
        \begin{aligned}
        \label{optimal_m}
            \mathbf{m}=\boldsymbol{\Psi}^{-1}\left(\boldsymbol{\gamma-\mathbf{P}^\mathrm{T}\boldsymbol{\alpha}}\right)
        \end{aligned}
    \end{align}
    Substituting equation~(\ref{optimal_m}) into equation~(\ref{convex_init_W}) yields
        \begin{align}
        \begin{aligned}
        \label{convex_init_W_only_alpha}
            E =& \boldsymbol{\alpha}^\mathrm{T}\left(\boldsymbol{\Theta}-\mathbf{P}\boldsymbol{\Psi}^{-1}\mathbf{P}^\mathrm{T}\right)\boldsymbol{\alpha}
                 -2\left(\boldsymbol{\xi}^\mathrm{T}-\boldsymbol{\gamma}^\mathrm{T}\boldsymbol{\Psi}^{-1}\mathbf{P}^\mathrm{T}\right)\boldsymbol{\alpha}
                 +\Phi-\boldsymbol{\gamma}^\mathrm{T}\boldsymbol{\Psi}^{-1}\boldsymbol{\gamma}
        \end{aligned}
    \end{align}
    Since equation~(\ref{convex_init_W_only_alpha}) is derived from the non-negative least square problem equation~(\ref{convex_init_W}), the quadratic in  equation~(\ref{convex_init_W_only_alpha}) must be positive semidefinite, and thus this problem is convex and admits a deterministic solution \cite{lawson1995solving, bro1997fast}.

    To impose non-negativity constraints on $\mathbf{S}$ and $\mathbf{Y}$ from equation~(\ref{optimal_m}) and equation~(\ref{convex_init_W_only_alpha}) without significantly increasing the computational demand, we use the truncation step of NNDSVD, which generates the best non-negative approximation of each new component\cite{boutsidis2008svd}.
    Since the truncation alters the entries in $\mathbf{S}$ and $\mathbf{Y}$, we let $\mathbf{W} = \left[\mathbf{W}_0\mathrm{diag}(\boldsymbol{\alpha})\ \mathbf{W}_\mathrm{new}\right]$ and $\mathbf{H} = \left[\mathbf{H}_0;\ \mathbf{H}_\mathrm{new}\right]$,  all components in $\mathbf{W}$ and $\mathbf{H}$ are re-optimized together by
    \begin{align}
        \label{final_amplitute_modification}
        \min_{\boldsymbol{\beta}\geq0}\lVert \mathbf{X}-\sum_{p=1}^{r_0+k}\beta_p\mathbf{w}_{p}\mathbf{h}_{p}^\mathrm{T}\rVert^2
    \end{align}
    where $\boldsymbol{\beta} = \left[\beta_1, \beta_2, \dots, \beta_{r_0+k}\right]$. $\mathbf{w}_{p}$, $\mathbf{h}_{p}$ are the $p$-th column and $p$-th row in $\mathbf{W}$, $\mathbf{H}$ respectively. 
    Equation~(\ref{final_amplitute_modification}) is also a nonnegative least squares problem and thus convex.
    The final step is refinement via NMF, initialized with $\mathbf{W}_{g} = \mathbf{W}\mathrm{diag}(\boldsymbol{\beta})$ and $\mathbf{H}_{g} = \mathbf{H}$. The GSVD-based feature recovery is summarized in Algorithm \ref{alg_gsvd}.

    If the SVD in equation~(\ref{max_obj_2}) has more components than the NMF, it is guaranteed that some singular values $\lambda$ will be infinite, and our strategy will first select directions missing from the NMF. When adopting this strategy, the schematic in Fig. \ref{fig_gsvd} depicting the polishing of existing directions is no longer relevant.
    
    \begin{algorithm}[H]
    \caption{GSVD-based feature recovery}\label{alg_gsvd}
    \begin{algorithmic}
        \STATE 
        \STATE \textbf{Input:} $\mathbf{W}_0,\ \mathbf{H}_0,\ \mathbf{U}_{r_0},\ \boldsymbol{\Sigma}_{r_0},\ \mathbf{V}_{r_0},\ k$
        \STATE \textbf{Output:} $\mathbf{W}_g,\ \mathbf{H}_{g}$
        \STATE \hspace{0.5cm} Compute $\mathbf{Q}$, $\mathbf{C}$, $\mathbf{G}$, $\mathbf{D}_1$, $\mathbf{D}_2$ by solving equation~(\ref{GSVD_solution}) and equation~(\ref{D_1andD_2})
        \STATE \hspace{0.5cm} $l \gets \mathrm{size}(\mathbf{G}, 1)$
        \STATE \hspace{0.5cm} $ \boldsymbol{\lambda}\gets\mathrm{vcat}\left(\mathrm{fill}(\mathrm{Inf}, r_0-l),\  \mathrm{diag}\left(\left(\mathbf{G}^\mathrm{T}\mathbf{G}\right)^\mathrm{-1}\mathbf{C}^\mathrm{T}\mathbf{C}\right)\right)$
        \STATE \hspace{0.5cm} $\mathbf{Y} \gets k$ columns in $\mathbf{V}_{r_0}\left(\mathbf{Q}^\mathrm{T}\right)^{-1}$ corresponding to the largest $k$ values in $\boldsymbol{\lambda}$ 
        \STATE \hspace{0.5cm} Compute $ \mathbf{S}, \boldsymbol{\alpha}$ with equation~(\ref{optimal_m}) and equation~(\ref{convex_init_W_only_alpha})
        \STATE \hspace{0.5cm} Compute $\mathbf{W}_\mathrm{new}$, $\mathbf{H}_\mathrm{new}$ by truncating $\mathbf{S}$, $\mathbf{Y}$ (Truncation step in NNDSVD) 
        \STATE \hspace{0.5cm} $\mathbf{W} \gets  \left[\mathbf{W}_0\mathrm{diag}(\boldsymbol{\alpha})\ \mathbf{W}_\mathrm{new}\right]$
        \STATE \hspace{0.5cm} $\mathbf{H} \gets \left[\mathbf{H}_0;\ \mathbf{H}_\mathrm{new}\right]$
        \STATE \hspace{0.5cm} Compute $\boldsymbol{\beta}$ with equation~(\ref{final_amplitute_modification})
        \STATE \hspace{0.5cm} $\mathbf{W}_g \gets \mathbf{W}\mathrm{diag}(\boldsymbol{\beta})$
        \STATE \hspace{0.5cm} $\mathbf{H}_g \gets \mathbf{H}$
    \end{algorithmic}
    \label{gsvd_alg}
\end{algorithm}

\section*{Results}

In this section, we demonstrate the effectiveness of GSVD-NMF and compare its performance with standard NMF on synthetic and real-world data. 
The simulation and experiments were performed using Julia 1.10 on Washington University in St. Louis RIS scientific computing platform with Intel\_Xeon\_Gold6242CPU280GHz 8G RAM.
On synthetic data, the NMF algorithm chosen for comparison and refinement in the GSVD-NMF pipeline Hierarchical alternating least squares (HALS)\cite{cichocki2009fast}, one of the most accurate and widely-used NMF algorithms.\cite{gillis2014and, gillis2012accelerated}.
On real-world data, in addition to HALS we also analyzed Greedy Coordinate Descent (GCD)\cite{hsieh2011fast}, Alternating Least Squares Using Projected Gradient Descent (ALSPGrad)\cite{lin2007projected}, and Multiplicative Updating (MU)\cite{lee2000algorithms}. After establishing efficacy with a variety of algorithms, later analysis on computational efficiency and influence of parameters is performed just with HALS.

To determine convergence, here we monitor the maximum relative change in the Frobenius norm of the columns of $\mathbf{W}\in\mathbb{R}^{m\times r}$ and the rows of $\mathbf{H}\in\mathbb{R}^{r\times n}$ (given in equation~(\ref{hals_stop_condition})), and stop iterating when
    \begin{align}
        \begin{aligned}
            \label{hals_stop_condition}
            \lVert\mathbf{w}_j^{(k+1)}-\mathbf{w}_j^{(k)}\rVert^2&\leq\epsilon\lVert\mathbf{w}_j^{(k+1)}+\mathbf{w}_j^{(k)}\rVert^2\\
            \lVert\mathbf{h}_j^{(k+1)}-\mathbf{h}_j^{(k)}\rVert^2&\leq\epsilon\lVert\mathbf{h}_j^{(k+1)}+\mathbf{h}_j^{(k)}\rVert^2
        \end{aligned}
    \end{align}
    for $j=1, 2, \dots, r$, for some choice of $\epsilon$. 
    In these expressions, $\mathbf{w}_j$ and $\mathbf{h}_j$ are the $j$-th column and $j$-th row in $\mathbf{W}$ and $\mathbf{H}$ respectively. 
     Unless otherwise specified, $\epsilon=10^{-4}$ was used. Throughout, the maximum number of iterations for NMF was set so high that termination was triggered only by the $\epsilon$-criterion.
    The relative fitting error between $\mathbf{W}\mathbf{H}$ and the input matrix $\mathbf{X}$, $100\lVert \mathbf{X}-\mathbf{W}\mathbf{H} \rVert_2^2/\lVert \mathbf{X} \rVert_2^2$, was used to evaluate the local optima of NMF. 
    This measures how well the factorization fit the input matrix.
    
\subsection*{Synthetic data: an illustrative example}

By identifying new directions from the SVD, GSVD-NMF has the potential to recover missing components from under-complete NMF results. 
To assess this, we first tested GSVD-NMF on synthetic data with 10 ground truth components, which are shown in Fig. \ref{simu_gsvd_k_1}(a).
Running HALS with 10 components, initialized using NNDSVD, on $\mathbf{W}\mathbf{H}$ with Gaussian noise (as shown in Fig. \ref{simu_gsvd_k_1}(b)) resulted in an inaccurate solution (Fig. \ref{simu_gsvd_k_1}(g)). The solution included multiple feature components and a noise component, with one feature failing to capture an independent component.

To exploit GSVD-NMF, we started from the solution returned by HALS with 9 components, one less than the number of ground truth components, obtaining the result shown in Fig. \ref{simu_gsvd_k_1}(c). 
One sees that several components of the ground truth are blended, Fig. \ref{simu_gsvd_k_1}(c).
To test for missing information, we compared the 9-component NMF with a 9-component SVD using GSVD as described in the ``Methods'' part. This identified a ``missing'' direction related to the first (large-magnitude) generalized singular value, Fig. \ref{simu_gsvd_k_1}(d). Note that this analysis provides support for only one additional component.
The component recovered by the feature recovery step in the pipeline (Fig. \ref{whole_pipe}) is shown in green in Fig. \ref{simu_gsvd_k_1}(e).
After using HALS to optimize this 10-component augmented factorization, 
the final NMF result, shown in Fig. \ref{simu_gsvd_k_1}(f), faithfully represents each ground truth component.
Therefore, GSVD-NMF can recover the missing component from under-complete NMF, and in this case performs better than naive HALS even when the correct number of components is known and a high-quality initialization strategy is used.

To test the performance of the GSVD-NMF more comprehensively, we tested the whole pipeline with independent random noise on each of 1000 trials. The comparison of relative fitting error between GSVD-NMF and naive HALS (initialized with NNDSVD) is shown in a scatter plot as Fig. \ref{simu_gsvd_k_1}(h). The diagonal represents equal performance, and the orthogonal brown line represents the histogram of perpendicular distances between points and the diagonal. The results in Fig. \ref{simu_gsvd_k_1}(h) suggests that GSVD-NMF outperforms or matches naive HALS in the large majority cases: on 857 out of 1000 trials, GSVD-NMF achieved a final objective value that was smaller or equal to that of naive HALS.

\subsection*{Experimental results}
\subsubsection*{Data sets}
\begin{figure*}[!t]
        \centering
        \includegraphics[width=5in]{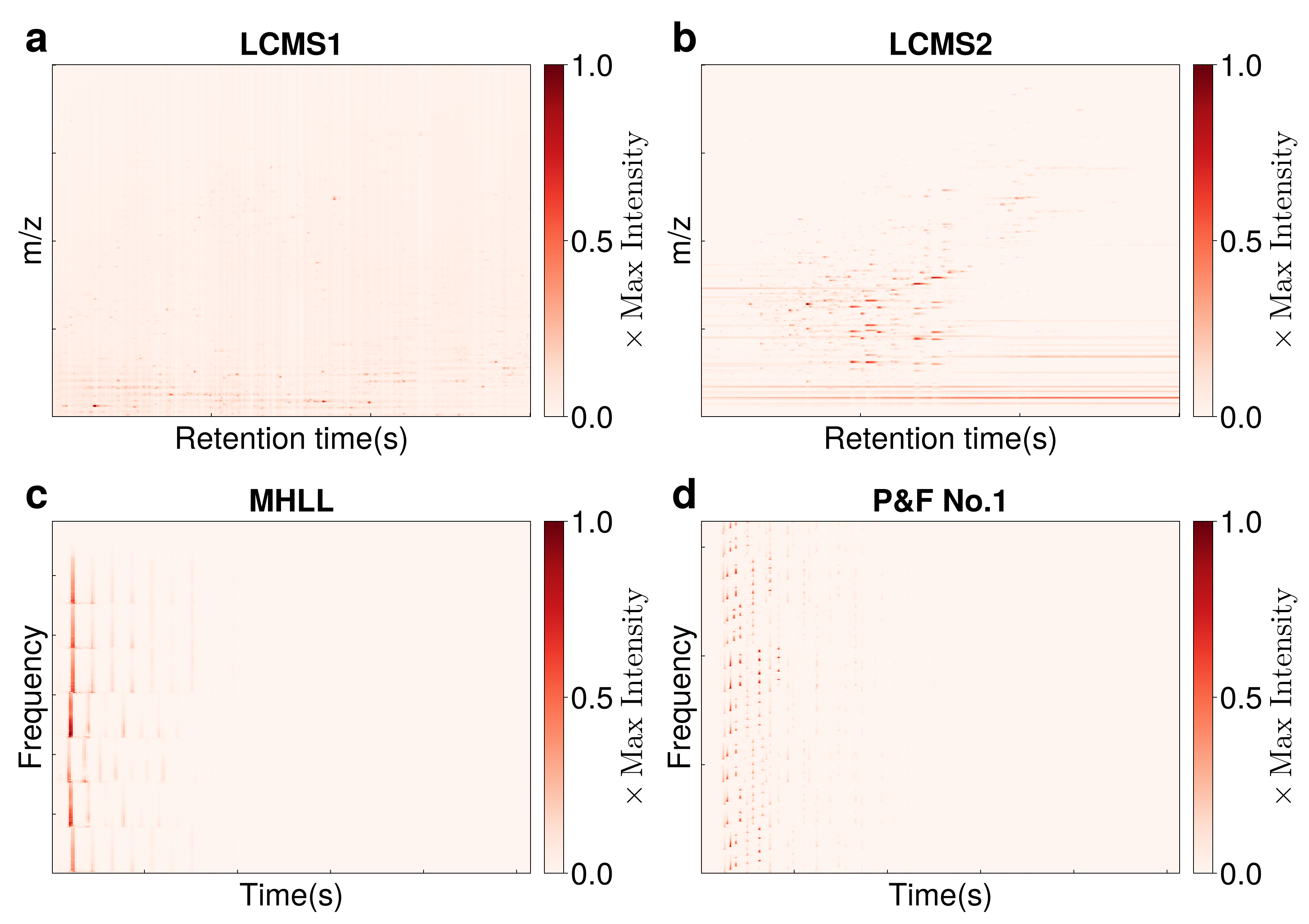}
        \caption{The real-world data sets used for experiments. (a) LCMS1. (b) LCMS2. (c) The amplitude spectrogram of "Mary had a little lamb". (d) The amplitude spectrogram of "Prelude and Fugue No.1 in C major". The colorbar label represents the intensity at each pixel, normalized by the maximum intensity of the matrix. For the LCMS data, the intensity corresponds to the ion count. Higher values indicate a greater number of ions.}
        \label{fig_datasets}
\end{figure*}

We also tested the performance of GSVD-NMF on four real-world datasets, which includes two liquid chromatography-mass spectrometry datasets (LCMS1 with doi:10.25345/C5KP7TV9T and LCMS2 with doi:10.25345/C58C9R77T), shown in Fig. \ref{fig_datasets}(a) and (b) respectively), and two audio datasets.
    The audio datasets feature the first measure of ``Mary had a little lamb'' and the first 30 seconds of ``Prelude and Fugue No.1 in C major'' by J.S. Bach, played by Glenn Gould, both of which are taken from reference \cite{leplat2020blind}.
    Fig. \ref{fig_datasets}(c) and (d) illustrates the amplitude spectrograms of audio data. 
    Details about all four real-world datasets are presented in Table~\ref{data_sets} and the column $r$ lists the ``real'' number of features we selected for this paper. 
    We choose the number of components for the four datasets as specified in Table~\ref{data_sets}. Where possible, these choices were based on previous literature with the same datasets\cite{leplat2020blind}.
    For the LCMS datasets not previously studied, we tested several rank-selection methods\cite{lin2020optimization, gillis2014and, tan2012automatic}. Because these methods gave divergent answers, ultimately we selected one among the recommended values based on visual inspection.
   
    \subsubsection*{Method of comparison}

    Since the ground truth components of real-world data are undetermined, we assessed methods in terms of the value of the objective equation~(\ref{SED_obj}), with smaller values indicating better solutions. 
    For each random initialization, we compared the final solution from GSVD-NMF against standard NMF with the same number of components.
    Standard NMF was initialized with a random rank-$r$ solution, with $r$ given in Table \ref{data_sets}.
    The under-complete NMF was initialized with the first $r_0 = r - k$ components of this same initialization, and then $k$ new components were added by GSVD-NMF to achieve rank $r$. 
    Thus, the two solutions start from as much of a shared initialization as can be achieved given their differences in initial rank.
    This comparison addresses whether it is more effective and efficient to directly augment components for an existing under-complete NMF solution or to rerun NMF from scratch with additional components.
    Solution objective values are compared in scatter plots, where each dot represents a single random initialization, and brown lines represent the histogram of perpendicular distances between points and the diagonal (a.k.a., equal performance) in the figure.
     
    \subsubsection*{GSVD-NMF recovers components for under-complete NMF}
    \label{GSVD-based method help NMF converge to better local optima}
    The comparison of local optima achieved by standard NMF and GSVD-NMF is given in Fig.~\ref{fig_scatters_spec_r_tol}. 
    Here, we show the results for $k=1$ (a single new component) and $k=0.2r_0$ (recovering $20\%$ of the number of components in the under-complete NMF).
    Both are evaluated via relative fitting error, with the 
    horizontal axis representing GSVD-NMF and the vertical axis standard NMF.

    Fig.~\ref{fig_scatters_spec_r_tol} demonstrates that NMF converges to single or multiple stationary points across all data sets using all four algorithms.
    Most are local minima, but for the MU algorithm some clusters in Fig.~\ref{fig_scatters_spec_r_tol}(d) contained solutions that could be substantially improved via HALS; since HALS optimizes equation~(\ref{SED_obj}) strictly by descent, these MU-solutions did not lie at the bottom of a local basin.
    For the alternating least squares algorithms (HALS, ALSPGrad, and GCD), most points in Fig.~\ref{fig_scatters_spec_r_tol}(a)-(c) are above or along the diagonal line, indicating that the recovered solution from under-complete NMF by GSVD-NMF either matches or improves the convergence of NMF in the majority of cases.
    This conclusion was true both when adding a single component ($k=1$) or when adding multiple components simultaneously ($k=0.2r_0$).
    In contrast, with the MU algorithm, GSVD-NMF did not consistently improve solution quality, and there were several instances where it was worse on average. 
    GSVD-NMF is therefore not recommended for use with MU, but given that MU itself was worse than all other tested alternatives, perhaps the best advice is to avoid use of MU.

    We also evaluated whether GSVD-NMF could achieve gains over existing approaches when using deterministic initialization methods such as NNDSVD, NNDSVDa, and NNDSVDar \cite{boutsidis2008svd}. 
    The results are shown in Table~\ref{diff_ini_compare}. 
    The NMF solutions recovered by GSVD-NMF are comparable to or better than standard NMF across all datasets when used with all three ALS-based algorithms, whether recovering single or multiple components simultaneously. 
    As with random initialization, GSVD-NMF was not an improvement for MU, which again performed worse than all three ALS algorithms,
    consistent with the results in Fig.~\ref{fig_scatters_spec_r_tol}(d).
    This outcome demonstrates the perhaps surprising point that GSVD-NMF can improve NMF even when the same SVD components have already been used to initialize NMF.

    \subsubsection*{Computational performance: GSVD-NMF finds high-quality solutions more quickly}
    Aside from the quality of solutions, many applications of NMF are sensitive to computational performance. 
    Superficially, one might imagine that GSVD-NMF would increase the computational cost, given the extra steps that include additional optimization runs.
    To investigate this issue, we measured the time required to augment an existing under-complete solution and compared it to the time required to run NMF again with more components.
    Specifically, starting from the same initialization we performed NMF twice: once to measure the average time per iteration (which varies with dataset and number of retained components), and a second time to measure the objective value after each iteration until convergence. Measuring these two quantities separately was necessary because computation of the objective value takes much longer than a single iteration of NMF.

    The results for random initialization are presented in Fig.\ref{fig_temporal_analysis}(a), as a scatter plot comparing the time required to augment components in under-complete NMF by GSVD-NMF versus the time to run NMF from scratch with additional components.
    The title of each panel within the subplot provides the name of the dataset.
    Similar to Fig.~\ref{fig_scatters_spec_r_tol}, each dot represents a single NMF run with a specific initialization (now measuring run-time rather than objective value), while the brown lines indicate the distribution of distances between each dot and the diagonal.
    Fig.~\ref{fig_temporal_analysis}(a) shows that in most cases, initializing new components by GSVD-NMF is generally more efficient than rerunning NMF with extra components. 
    As suggested by the thin vertical clusters that frequently appear in these plots, the time taken for GSVD-NMF augmentation was relatively consistent for each under-complete local minimum, while the times for running NMF from scratch with additional random components varied significantly.

    For deterministic initializations (SVD-based initialization), Fig.~\ref{fig_temporal_analysis}(b)-(d) shows the traces of objective value versus time for different initialization methods across four datasets. Each panel’s caption includes the dataset name and the number of new components added.
    The GSVD-NMF trace represents the final NMF phase used for refinement. 
    The first element in each trace represents the objective value after one iteration. 
    Fig.~\ref{fig_temporal_analysis}(b)-(d) indicates that the final NMF in GSVD-NMF converges faster than standard NMF for both LCMS datasets and ``Mary Had a Little Lamb.''
    In contrast, for ``Prelude and Fugue No.1 in C major,'' the convergence performance is similar to or slightly worse than that of standard NMF.
    Further zooming into the region around the intersection of the two strategies’ traces reveals that for $k=1$, the proposed method behaves similarly to standard NMF. However, for $k=2$, it achieves a slightly lower objective value after a period of slow progress, indicating a minor advantage in the final objective, albeit with slower convergence.

    In conclusion, we find that it is generally more efficient to augment a prior solution by GSVD-NMF than it is to run standard NMF from scratch with additional components.
    Thus, on average, GSVD-NMF improves upon both quality and computational performance.

    The performance of each stage in GSVD-NMF is documented in Table~\ref{time_of_each_part}. It is evident that the `Feature Recovery' step is highly efficient, representing a nearly negligible portion of the total time across all datasets. Therefore, the efficiency of GSVD-NMF primarily depends on the time allocated to the NMF step in the pipeline. In other words, $\epsilon$ (or the maximum number of iterations) serves as the main control parameter for the runtime performance of GSVD-NMF.

    \subsubsection*{The choice of $\epsilon_0$ (utilizing GSVD-NMF as a general NMF pipeline)}
    
    Given its success, GSVD-NMF can be recommended as a general NMF pipeline. For such uses, there is one additional potential optimization worthy of investigation: GSVD-NMF runs standard NMF twice, once for the initial under-complete NMF and again for the final NMF (Fig.~\ref{whole_pipe}).
    So far, we employed the same $\epsilon = 10^{-4}$ in equation~(\ref{hals_stop_condition}) for both NMF runs.
    Here, we investigate whether one can reduce computation time by relaxing the tolerance of the initial under-complete NMF without sacrificing overall quality.

    The effects on solution quality of reducing the stringency of the under-complete NMF ($\epsilon_0$) steps are shown in Fig.~\ref{fig_scatters_spec_r_all_tol}. 
    Here, we present the results from HALS for NMF with $k=1$. The tolerance had little effect on the ``Mary had a little lamb'' dataset. For the other three datasets, the advantages of GSVD-NMF increase with the stringency of $\epsilon_0$, particularly evident in the ``Prelude and Fugue No.1 in C major'' dataset.
    Overall, these results demonstrate that in GSVD-NMF, the under-complete NMF can be run at higher tolerance (lower stringency) but with higher risk of landing in worse local optima. However, at all tested stringencies GSVD-NMF outperforms or matches standard NMF in aggregate. 
    Therefore, based on the results of this section and Fig.~\ref{fig_scatters_spec_r_tol}, GSVD-NMF could be considered a general pipeline for performing NMF.

 \subsubsection*{The choice of $k$}
    It should be noted that all of the above experiments set $k=1$ or $k=0.2r_0$, both of which appear to be broadly successful.
    In practice, we may wish to add more components simultaneously, rather than iteratively adding one component at a time, to make the whole pipeline more efficient. 
    Thus, it is valuable to investigate the role of $k$ on the final results.
    
\begin{figure}[!t]
        \centering
        \includegraphics[width=6in]{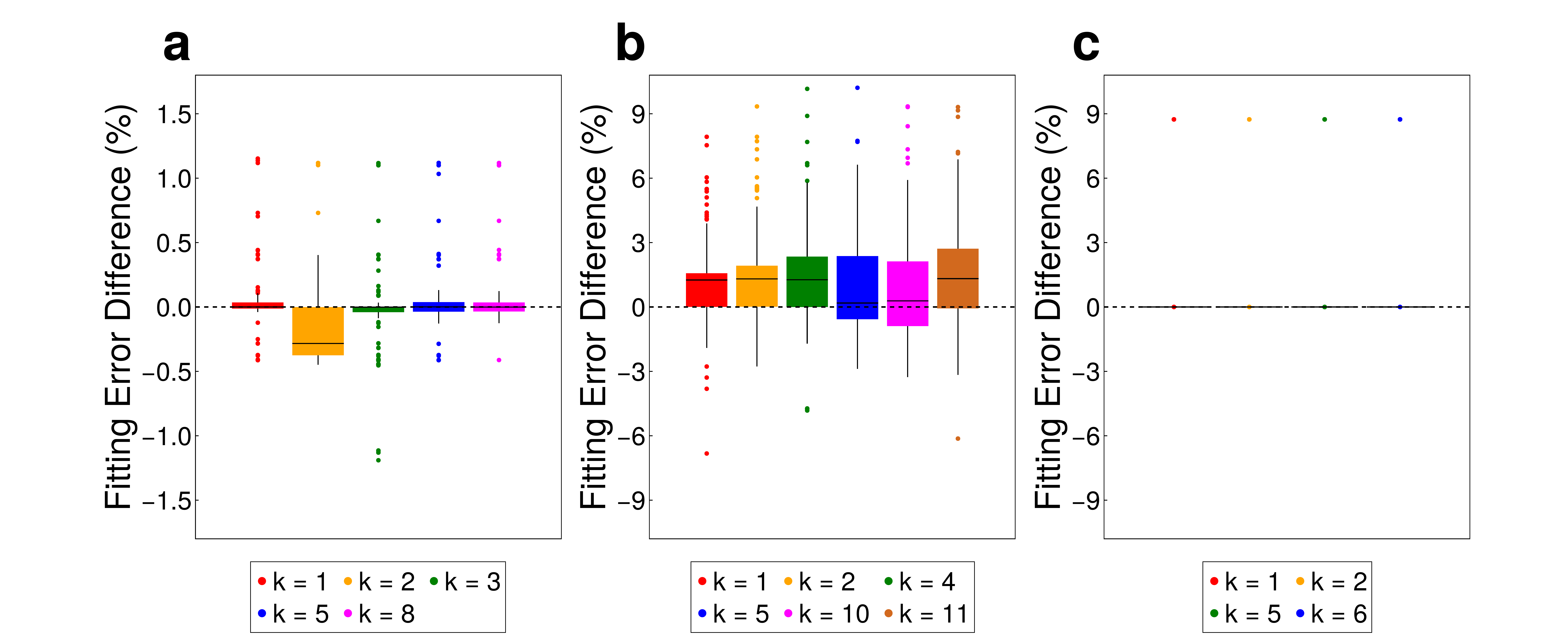}
        \caption{The effect of different $k$: (a) LCMS1. (b) LCMS2. (c) Prelude and Fugue No.1 in C major.
    This figure compares the difference in relative fitting error between GSVD-NMF and standard NMF ($\mathrm{Error}_\mathrm{Standard NMF}$-$\mathrm{Error}_\mathrm{GSVD-NMF}$) across different numbers of augmented components. The box plot summarizes the distribution of fitting error differences for each case. Positive values indicate that GSVD-NMF achieves a better fit, while negative values suggest that standard NMF performs better.
    }
        \label{box_plot_kadd}
    \end{figure}

    The experiment for Fig.~\ref{fig_scatters_spec_r_tol} (200 random initialized trials) was conducted with different $k$.   
    Fig.~\ref{box_plot_kadd} illustrates the difference between the fitting error of standard NMF and that of GSVD-NMF. 
    Consistent with the results in Fig.~\ref{fig_scatters_spec_r_tol}, GSVD-NMF finds superior local optimum in most cases on both LCMS data sets and ``Prelude and Fugue No.1 in C major'' regardless of $k$.
    Thus, the choice of $k$ is not critical, as a wide range of choices (at least for fairly small values of $k$) work well.
 
\section*{Conclusion}

    In conclusion, we propose a new strategy to reconstruct missing components from under-complete or poor local NMF solutions. Our strategy exploits the generalized SVD between the initial under-complete NMF solution and the SVD of the same rank to identify ``missing features'' and suggest new component(s), which are then refined by another round of NMF.
    We show that GSVD-NMF is capable of iteratively recovering missing components for under-complete NMF for several different well-established NMF algorithms, enabling a strategy in which one incrementally grows the rank of an NMF solution without the need to start from scratch. This may have practical benefits in real applications, particularly for large data sets, when the number of components is not known.
    Even when compared against NMF of the same rank, GSVD-NMF often outperformed standard NMF in terms of the quality of its local optima and the efficiency of the whole pipeline.


\begin{thebibliography}{10}
\urlstyle{rm}
\expandafter\ifx\csname url\endcsname\relax
  \def\url#1{\texttt{#1}}\fi
\expandafter\ifx\csname urlprefix\endcsname\relax\def\urlprefix{URL }\fi
\expandafter\ifx\csname doiprefix\endcsname\relax\def\doiprefix{DOI: }\fi
\providecommand{\bibinfo}[2]{#2}
\providecommand{\eprint}[2][]{\url{#2}}

\bibitem{lee1999learning}
\bibinfo{author}{Lee, D.~D.} \& \bibinfo{author}{Seung, H.~S.}
\newblock \bibinfo{journal}{\bibinfo{title}{Learning the parts of objects by non-negative matrix factorization}}.
\newblock {\emph{\JournalTitle{nature}}} \textbf{\bibinfo{volume}{401}}, \bibinfo{pages}{788--791} (\bibinfo{year}{1999}).

\bibitem{rajabi2014spectral}
\bibinfo{author}{Rajabi, R.} \& \bibinfo{author}{Ghassemian, H.}
\newblock \bibinfo{journal}{\bibinfo{title}{Spectral unmixing of hyperspectral imagery using multilayer nmf}}.
\newblock {\emph{\JournalTitle{IEEE Geoscience and Remote Sensing Letters}}} \textbf{\bibinfo{volume}{12}}, \bibinfo{pages}{38--42} (\bibinfo{year}{2014}).

\bibitem{wang2016hypergraph}
\bibinfo{author}{Wang, W.}, \bibinfo{author}{Qian, Y.} \& \bibinfo{author}{Tang, Y.~Y.}
\newblock \bibinfo{journal}{\bibinfo{title}{Hypergraph-regularized sparse nmf for hyperspectral unmixing}}.
\newblock {\emph{\JournalTitle{IEEE journal of selected topics in applied earth observations and remote sensing}}} \textbf{\bibinfo{volume}{9}}, \bibinfo{pages}{681--694} (\bibinfo{year}{2016}).

\bibitem{chen2019experimental}
\bibinfo{author}{Chen, Y.}, \bibinfo{author}{Zhang, H.}, \bibinfo{author}{Liu, R.}, \bibinfo{author}{Ye, Z.} \& \bibinfo{author}{Lin, J.}
\newblock \bibinfo{journal}{\bibinfo{title}{Experimental explorations on short text topic mining between lda and nmf based schemes}}.
\newblock {\emph{\JournalTitle{Knowledge-Based Systems}}} \textbf{\bibinfo{volume}{163}}, \bibinfo{pages}{1--13} (\bibinfo{year}{2019}).

\bibitem{kowsari2019text}
\bibinfo{author}{Kowsari, K.} \emph{et~al.}
\newblock \bibinfo{journal}{\bibinfo{title}{Text classification algorithms: A survey}}.
\newblock {\emph{\JournalTitle{Information}}} \textbf{\bibinfo{volume}{10}}, \bibinfo{pages}{150} (\bibinfo{year}{2019}).

\bibitem{gemmeke2013exemplar}
\bibinfo{author}{Gemmeke, J.~F.}, \bibinfo{author}{Vuegen, L.}, \bibinfo{author}{Karsmakers, P.}, \bibinfo{author}{Vanrumste, B.} \emph{et~al.}
\newblock \bibinfo{title}{An exemplar-based nmf approach to audio event detection}.
\newblock In \emph{\bibinfo{booktitle}{2013 IEEE workshop on applications of signal processing to audio and acoustics}}, \bibinfo{pages}{1--4} (\bibinfo{organization}{IEEE}, \bibinfo{year}{2013}).

\bibitem{leplat2020blind}
\bibinfo{author}{Leplat, V.}, \bibinfo{author}{Gillis, N.} \& \bibinfo{author}{Ang, A.~M.}
\newblock \bibinfo{journal}{\bibinfo{title}{Blind audio source separation with minimum-volume beta-divergence nmf}}.
\newblock {\emph{\JournalTitle{IEEE Transactions on Signal Processing}}} \textbf{\bibinfo{volume}{68}}, \bibinfo{pages}{3400--3410} (\bibinfo{year}{2020}).

\bibitem{dubroca2012weighted}
\bibinfo{author}{Dubroca, R.}, \bibinfo{author}{Junor, C.} \& \bibinfo{author}{Souloumiac, A.}
\newblock \bibinfo{title}{Weighted nmf for high-resolution mass spectrometry analysis}.
\newblock In \emph{\bibinfo{booktitle}{2012 Proceedings of the 20th European Signal Processing Conference (EUSIPCO)}}, \bibinfo{pages}{1806--1810} (\bibinfo{organization}{IEEE}, \bibinfo{year}{2012}).

\bibitem{lin2020optimization}
\bibinfo{author}{Lin, X.} \& \bibinfo{author}{Boutros, P.~C.}
\newblock \bibinfo{journal}{\bibinfo{title}{Optimization and expansion of non-negative matrix factorization}}.
\newblock {\emph{\JournalTitle{BMC bioinformatics}}} \textbf{\bibinfo{volume}{21}}, \bibinfo{pages}{7} (\bibinfo{year}{2020}).

\bibitem{mcclure2014novel}
\bibinfo{author}{McClure, P.} \emph{et~al.}
\newblock \bibinfo{journal}{\bibinfo{title}{A novel nmf guided level-set for dwi prostate segmentation}}.
\newblock {\emph{\JournalTitle{Journal of Computer Science \& Systems Biology}}} \textbf{\bibinfo{volume}{7}}, \bibinfo{pages}{1} (\bibinfo{year}{2014}).

\bibitem{cichocki2009fast}
\bibinfo{author}{Cichocki, A.} \& \bibinfo{author}{Phan, A.-H.}
\newblock \bibinfo{journal}{\bibinfo{title}{Fast local algorithms for large scale nonnegative matrix and tensor factorizations}}.
\newblock {\emph{\JournalTitle{IEICE transactions on fundamentals of electronics, communications and computer sciences}}} \textbf{\bibinfo{volume}{92}}, \bibinfo{pages}{708--721} (\bibinfo{year}{2009}).

\bibitem{lin2007projected}
\bibinfo{author}{Lin, C.-J.}
\newblock \bibinfo{journal}{\bibinfo{title}{Projected gradient methods for nonnegative matrix factorization}}.
\newblock {\emph{\JournalTitle{Neural computation}}} \textbf{\bibinfo{volume}{19}}, \bibinfo{pages}{2756--2779} (\bibinfo{year}{2007}).

\bibitem{gillis2010using}
\bibinfo{author}{Gillis, N.} \& \bibinfo{author}{Glineur, F.}
\newblock \bibinfo{journal}{\bibinfo{title}{Using underapproximations for sparse nonnegative matrix factorization}}.
\newblock {\emph{\JournalTitle{Pattern recognition}}} \textbf{\bibinfo{volume}{43}}, \bibinfo{pages}{1676--1687} (\bibinfo{year}{2010}).

\bibitem{gillis2014and}
\bibinfo{author}{Gillis, N.}
\newblock \bibinfo{journal}{\bibinfo{title}{The why and how of nonnegative matrix factorization}}.
\newblock {\emph{\JournalTitle{Regularization, optimization, kernels, and support vector machines}}} \textbf{\bibinfo{volume}{12}}, \bibinfo{pages}{257--291} (\bibinfo{year}{2014}).

\bibitem{fogel2023rank}
\bibinfo{author}{Fogel, P.}, \bibinfo{author}{Geissler, C.}, \bibinfo{author}{Morizet, N.} \& \bibinfo{author}{Luta, G.}
\newblock \bibinfo{journal}{\bibinfo{title}{On rank selection in non-negative matrix factorization using concordance}}.
\newblock {\emph{\JournalTitle{Mathematics}}} \textbf{\bibinfo{volume}{11}}, \bibinfo{pages}{4611} (\bibinfo{year}{2023}).

\bibitem{biggs2008nonnegative}
\bibinfo{author}{Biggs, M.}, \bibinfo{author}{Ghodsi, A.} \& \bibinfo{author}{Vavasis, S.}
\newblock \bibinfo{title}{Nonnegative matrix factorization via rank-one downdate}.
\newblock In \emph{\bibinfo{booktitle}{Proceedings of the 25th International Conference on Machine learning}}, \bibinfo{pages}{64--71} (\bibinfo{year}{2008}).

\bibitem{kim2014algorithms}
\bibinfo{author}{Kim, J.}, \bibinfo{author}{He, Y.} \& \bibinfo{author}{Park, H.}
\newblock \bibinfo{journal}{\bibinfo{title}{Algorithms for nonnegative matrix and tensor factorizations: A unified view based on block coordinate descent framework}}.
\newblock {\emph{\JournalTitle{Journal of Global Optimization}}} \textbf{\bibinfo{volume}{58}}, \bibinfo{pages}{285--319} (\bibinfo{year}{2014}).

\bibitem{hoyer2004non}
\bibinfo{author}{Hoyer, P.~O.}
\newblock \bibinfo{journal}{\bibinfo{title}{Non-negative matrix factorization with sparseness constraints.}}
\newblock {\emph{\JournalTitle{Journal of machine learning research}}} \textbf{\bibinfo{volume}{5}} (\bibinfo{year}{2004}).

\bibitem{morup2009tuning}
\bibinfo{author}{M{\o}rup, M.} \& \bibinfo{author}{Hansen, L.~K.}
\newblock \bibinfo{title}{Tuning pruning in sparse non-negative matrix factorization}.
\newblock In \emph{\bibinfo{booktitle}{2009 17th European Signal Processing Conference}}, \bibinfo{pages}{1923--1927} (\bibinfo{organization}{IEEE}, \bibinfo{year}{2009}).

\bibitem{hinrich2018probabilistic}
\bibinfo{author}{Hinrich, J.~L.} \& \bibinfo{author}{M{\o}rup, M.}
\newblock \bibinfo{title}{Probabilistic sparse non-negative matrix factorization}.
\newblock In \emph{\bibinfo{booktitle}{Latent Variable Analysis and Signal Separation: 14th International Conference, LVA/ICA 2018, Guildford, UK, July 2--5, 2018, Proceedings 14}}, \bibinfo{pages}{488--498} (\bibinfo{organization}{Springer}, \bibinfo{year}{2018}).

\bibitem{li2013non}
\bibinfo{author}{Li, Y.} \& \bibinfo{author}{Ngom, A.}
\newblock \bibinfo{journal}{\bibinfo{title}{The non-negative matrix factorization toolbox for biological data mining}}.
\newblock {\emph{\JournalTitle{Source code for biology and medicine}}} \textbf{\bibinfo{volume}{8}}, \bibinfo{pages}{1--15} (\bibinfo{year}{2013}).

\bibitem{arora2012computing}
\bibinfo{author}{Arora, S.}, \bibinfo{author}{Ge, R.}, \bibinfo{author}{Kannan, R.} \& \bibinfo{author}{Moitra, A.}
\newblock \bibinfo{title}{Computing a nonnegative matrix factorization--provably}.
\newblock In \emph{\bibinfo{booktitle}{Proceedings of the forty-fourth annual ACM symposium on Theory of computing}}, \bibinfo{pages}{145--162} (\bibinfo{year}{2012}).

\bibitem{boutsidis2008svd}
\bibinfo{author}{Boutsidis, C.} \& \bibinfo{author}{Gallopoulos, E.}
\newblock \bibinfo{journal}{\bibinfo{title}{Svd based initialization: A head start for nonnegative matrix factorization}}.
\newblock {\emph{\JournalTitle{Pattern recognition}}} \textbf{\bibinfo{volume}{41}}, \bibinfo{pages}{1350--1362} (\bibinfo{year}{2008}).

\bibitem{esposito2021review}
\bibinfo{author}{Esposito, F.}
\newblock \bibinfo{journal}{\bibinfo{title}{A review on initialization methods for nonnegative matrix factorization: towards omics data experiments}}.
\newblock {\emph{\JournalTitle{Mathematics}}} \textbf{\bibinfo{volume}{9}}, \bibinfo{pages}{1006} (\bibinfo{year}{2021}).

\bibitem{fathi2023initialization}
\bibinfo{author}{Fathi~Hafshejani, S.} \& \bibinfo{author}{Moaberfard, Z.}
\newblock \bibinfo{journal}{\bibinfo{title}{Initialization for non-negative matrix factorization: a comprehensive review}}.
\newblock {\emph{\JournalTitle{International Journal of Data Science and Analytics}}} \textbf{\bibinfo{volume}{16}}, \bibinfo{pages}{119--134} (\bibinfo{year}{2023}).

\bibitem{casalino2014subtractive}
\bibinfo{author}{Casalino, G.}, \bibinfo{author}{Del~Buono, N.} \& \bibinfo{author}{Mencar, C.}
\newblock \bibinfo{journal}{\bibinfo{title}{Subtractive clustering for seeding non-negative matrix factorizations}}.
\newblock {\emph{\JournalTitle{Information Sciences}}} \textbf{\bibinfo{volume}{257}}, \bibinfo{pages}{369--387} (\bibinfo{year}{2014}).

\bibitem{alshabrawy2012underdetermined}
\bibinfo{author}{Alshabrawy, O.~S.}, \bibinfo{author}{Ghoneim, M.}, \bibinfo{author}{Awad, W.} \& \bibinfo{author}{Hassanien, A.~E.}
\newblock \bibinfo{title}{Underdetermined blind source separation based on fuzzy c-means and semi-nonnegative matrix factorization}.
\newblock In \emph{\bibinfo{booktitle}{2012 Federated Conference on Computer Science and Information Systems (FedCSIS)}}, \bibinfo{pages}{695--700} (\bibinfo{organization}{IEEE}, \bibinfo{year}{2012}).

\bibitem{atif2019improved}
\bibinfo{author}{Atif, S.~M.}, \bibinfo{author}{Qazi, S.} \& \bibinfo{author}{Gillis, N.}
\newblock \bibinfo{journal}{\bibinfo{title}{Improved svd-based initialization for nonnegative matrix factorization using low-rank correction}}.
\newblock {\emph{\JournalTitle{Pattern Recognition Letters}}} \textbf{\bibinfo{volume}{122}}, \bibinfo{pages}{53--59} (\bibinfo{year}{2019}).

\bibitem{yuan2020convex}
\bibinfo{author}{Yuan, A.}, \bibinfo{author}{You, M.}, \bibinfo{author}{He, D.} \& \bibinfo{author}{Li, X.}
\newblock \bibinfo{journal}{\bibinfo{title}{Convex non-negative matrix factorization with adaptive graph for unsupervised feature selection}}.
\newblock {\emph{\JournalTitle{IEEE Transactions on cybernetics}}} \textbf{\bibinfo{volume}{52}}, \bibinfo{pages}{5522--5534} (\bibinfo{year}{2020}).

\bibitem{huang2020robust}
\bibinfo{author}{Huang, Q.}, \bibinfo{author}{Yin, X.}, \bibinfo{author}{Chen, S.}, \bibinfo{author}{Wang, Y.} \& \bibinfo{author}{Chen, B.}
\newblock \bibinfo{journal}{\bibinfo{title}{Robust nonnegative matrix factorization with structure regularization}}.
\newblock {\emph{\JournalTitle{Neurocomputing}}} \textbf{\bibinfo{volume}{412}}, \bibinfo{pages}{72--90} (\bibinfo{year}{2020}).

\bibitem{ince2022weighted}
\bibinfo{author}{Ince, T.} \& \bibinfo{author}{Dobigeon, N.}
\newblock \bibinfo{journal}{\bibinfo{title}{Weighted residual nmf with spatial regularization for hyperspectral unmixing}}.
\newblock {\emph{\JournalTitle{IEEE Geoscience and Remote Sensing Letters}}} \textbf{\bibinfo{volume}{19}}, \bibinfo{pages}{1--5} (\bibinfo{year}{2022}).

\bibitem{jiao2020hyper}
\bibinfo{author}{Jiao, C.-N.}, \bibinfo{author}{Gao, Y.-L.}, \bibinfo{author}{Yu, N.}, \bibinfo{author}{Liu, J.-X.} \& \bibinfo{author}{Qi, L.-Y.}
\newblock \bibinfo{journal}{\bibinfo{title}{Hyper-graph regularized constrained nmf for selecting differentially expressed genes and tumor classification}}.
\newblock {\emph{\JournalTitle{IEEE journal of biomedical and health informatics}}} \textbf{\bibinfo{volume}{24}}, \bibinfo{pages}{3002--3011} (\bibinfo{year}{2020}).

\bibitem{deng2022graph}
\bibinfo{author}{Deng, P.} \emph{et~al.}
\newblock \bibinfo{journal}{\bibinfo{title}{Graph regularized sparse non-negative matrix factorization for clustering}}.
\newblock {\emph{\JournalTitle{IEEE Transactions on Computational Social Systems}}}  (\bibinfo{year}{2022}).

\bibitem{callaerts1990comparison}
\bibinfo{author}{Callaerts, D.} \emph{et~al.}
\newblock \bibinfo{journal}{\bibinfo{title}{Comparison of svd methods to extract the foetal electrocardiogram from cutaneous electrode signals}}.
\newblock {\emph{\JournalTitle{Medical and Biological Engineering and Computing}}} \textbf{\bibinfo{volume}{28}}, \bibinfo{pages}{217--224} (\bibinfo{year}{1990}).

\bibitem{guo2024optimal}
\bibinfo{author}{Guo, Y.} \& \bibinfo{author}{Holy, T.~E.}
\newblock \bibinfo{journal}{\bibinfo{title}{An optimal pairwise merge algorithm improves the quality and consistency of nonnegative matrix factorization}}.
\newblock {\emph{\JournalTitle{arXiv preprint arXiv:2408.09013}}}  (\bibinfo{year}{2024}).

\bibitem{li2015rayleigh}
\bibinfo{author}{Li, R.-C.}
\newblock \bibinfo{title}{Rayleigh quotient based optimization methods for eigenvalue problems}.
\newblock In \emph{\bibinfo{booktitle}{Matrix Functions and Matrix Equations}}, \bibinfo{pages}{76--108} (\bibinfo{publisher}{World Scientific}, \bibinfo{year}{2015}).

\bibitem{lawson1995solving}
\bibinfo{author}{Lawson, C.~L.} \& \bibinfo{author}{Hanson, R.~J.}
\newblock \emph{\bibinfo{title}{Solving least squares problems}} (\bibinfo{publisher}{SIAM}, \bibinfo{year}{1995}).

\bibitem{bro1997fast}
\bibinfo{author}{Bro, R.} \& \bibinfo{author}{De~Jong, S.}
\newblock \bibinfo{journal}{\bibinfo{title}{A fast non-negativity-constrained least squares algorithm}}.
\newblock {\emph{\JournalTitle{Journal of Chemometrics: A Journal of the Chemometrics Society}}} \textbf{\bibinfo{volume}{11}}, \bibinfo{pages}{393--401} (\bibinfo{year}{1997}).

\bibitem{gillis2012accelerated}
\bibinfo{author}{Gillis, N.} \& \bibinfo{author}{Glineur, F.}
\newblock \bibinfo{journal}{\bibinfo{title}{Accelerated multiplicative updates and hierarchical als algorithms for nonnegative matrix factorization}}.
\newblock {\emph{\JournalTitle{Neural computation}}} \textbf{\bibinfo{volume}{24}}, \bibinfo{pages}{1085--1105} (\bibinfo{year}{2012}).

\bibitem{hsieh2011fast}
\bibinfo{author}{Hsieh, C.-J.} \& \bibinfo{author}{Dhillon, I.~S.}
\newblock \bibinfo{title}{Fast coordinate descent methods with variable selection for non-negative matrix factorization}.
\newblock In \emph{\bibinfo{booktitle}{Proceedings of the 17th ACM SIGKDD international conference on Knowledge discovery and data mining}}, \bibinfo{pages}{1064--1072} (\bibinfo{year}{2011}).

\bibitem{lee2000algorithms}
\bibinfo{author}{Lee, D.} \& \bibinfo{author}{Seung, H.~S.}
\newblock \bibinfo{journal}{\bibinfo{title}{Algorithms for non-negative matrix factorization}}.
\newblock {\emph{\JournalTitle{Advances in neural information processing systems}}} \textbf{\bibinfo{volume}{13}} (\bibinfo{year}{2000}).

\bibitem{tan2012automatic}
\bibinfo{author}{Tan, V.~Y.} \& \bibinfo{author}{F{\'e}votte, C.}
\newblock \bibinfo{journal}{\bibinfo{title}{Automatic relevance determination in nonnegative matrix factorization with the/spl beta/-divergence}}.
\newblock {\emph{\JournalTitle{IEEE transactions on pattern analysis and machine intelligence}}} \textbf{\bibinfo{volume}{35}}, \bibinfo{pages}{1592--1605} (\bibinfo{year}{2012}).

\end{thebibliography}

\section*{Acknowledgements}
This work was supported by NIH grants R01DC020034 and R01DC010381 (T.E.H.), and training grant T32EB014855 (Y.G., PI: Joseph P. Culver).

\section*{Author contributions statement}
Y.G and T.E.H developed the method, Y.G conducted the experiments. Y.G and T.E.H analyzed the results. All authors reviewed the manuscript. 

\section*{Additional information}

\subsection*{Competing interests} 
The authors declare no competing interests.

\subsection*{Accession codes}
All code required for this paper is open-source and available at https://github.com/HolyLab/GsvdInitialization.jl

\section*{Legends}

\begin{figure*}[!t]
    \centering
    \includegraphics[width=6.9in]{images/simu_gt_standard_nmf.pdf}
    \caption{A synthetic example used to illustrate the GSVD-NMF for feature recovery ($k=1$), displaying $\mathbf{W}$ and $\mathbf{H}$. (a) Ground truth $\mathbf{W}$ (each line depicting one column) and $\mathbf{H}$ (each line depicting one row) with 10 features. (b) $\mathbf{X}$ generated as $\mathbf{W}\mathbf{H}$ with added Gaussian noise (c) Standard NMF results (HALS) with 9 components. (d) The generalized singular value spectrum from equation (\ref{gen_eigen_final}). (e) Feature recovery results ($\mathbf{W}_g$, $\mathbf{H}_g$), with the new component in green. (f) Final NMF results ($\mathbf{W}_1$, $\mathbf{H}_1$). (g) Standard NMF results initialized with NNDSVD. Despite knowing the correct number of components, several features are incompletely separated, and the solution is much worse than panel (f). (h) Comparing the fitting error of standard NMF and GSVD-NMF with 1000 trials of adding random Gaussian noise to $\mathbf{W}\mathbf{H}$ (initialization using NNDSVD).  
    The scatter plot compares the relative fitting errors of standard NMF and GSVD-NMF against the original matrix. Each magenta scatter point represents an individual comparison from a single random Gaussian noise added to $\mathbf{WH}$, with its position indicating the relative fitting error for standard NMF (vertical axis) and GSVD-NMF (horizontal axis). The brown line illustrates the histogram of the perpendicular distances from the scatter points to the diagonal, summarizing the overall distribution of error differences. For most tests, GSVD-NMF produces an equal or better fit.}
    \label{simu_gsvd_k_1}
\end{figure*}

    \begin{figure*}[!t]
        \centering
        \includegraphics[width=6.93in]{images/fig_scatters_spec_r_tol_multi_alg_multi_k.pdf}
        \caption{Comparing the fitting error of standard NMF and GSVD-NMF on real-world data. Each column corresponds to a different data set and/or number of components recovered by GSVD-NMF. (a) HALS. (b) GCD. (c) ALSGrad. (d) MU. Note that the axes for MU are expanded compared to the other three algorithms. Similar to the comparison shown in Fig.~\ref{simu_gsvd_k_1}(h), 
        the scatter plot compares the relative fitting errors of standard NMF and GSVD-NMF against the original matrix across different NMF algorithms. The key difference is that each magenta scatter point here represents an individual comparison from a single random initialization of NMF. The brown line illustrates the histogram of the perpendicular distances from the scatter points to the diagonal, summarizing the overall distribution of error differences.}
        \label{fig_scatters_spec_r_tol}
    \end{figure*} 

    \begin{figure*}[!t]
        \centering
        \includegraphics[width=6.93in]{images/fig_temporal_analysis.pdf}
        \caption{Runtimes of standard NMF vs GSVD-NMF with different initializations. (a) Total runtimes of the two algorithms with different random initializations. Each point represents the runtime for standard NMF plotted against the runtime for GSVD-NMF; points above the diagonal indicate shorter runtime for GSVD-NMF. The brown line shows the histogram of the perpendicular distances from the diagonal, summarizing the overall distribution of time differences. The remaining plots show detailed convergence trajectories (objective value vs.\ time) during iteration for deterministic initialization. (b) NNDSVD. (c) NNDSVDa. (d) NNDSVDar. For most data sets and initializations, GSVD-NMF converged to high-quality solutions more quickly.}
        \label{fig_temporal_analysis}
    \end{figure*} 

   \begin{figure*}[!t]
        \centering
        \includegraphics[width=6.93in]{images/fig_scatters_spec_r_all_tol.pdf}
        \caption{The effect of convergence tolerance $\epsilon_0$ on final results of GSVD-NMF. Panels should be compared to those of Fig. \ref{fig_scatters_spec_r_tol}, which used $\epsilon_0 = 10^{-4}$. (a) LCMS1. (b) LCMS2. (c) Mary had a little lamb. (d) Prelude and Fugue No.1 in C major.
        }
        \label{fig_scatters_spec_r_all_tol}
    \end{figure*}

    \clearpage
    \begin{table}[!t]
    \renewcommand{\arraystretch}{1.6}
        \centering
        \begin{tabular}{|c|M{7cm}|c|c|} \hline 
             No.& Data sets & Size & $r$\\ \hline 
             1& LCMS1 & $600\times400$ & 17\\ \hline
             2& LCMS2 & $600\times400$ & 23\\ \hline
             3& Mary had a little lamb (MHLL) & $294\times257$ & 3\\ \hline
             4& Prelude and Fugue No.1 in C major (P\&F No.1) & $647\times513$ & 13 \\
             \hline
        \end{tabular}
        \caption{Description of data sets}
        \label{data_sets}
    \end{table}
    
    \begin{table*}[!t]
    \renewcommand{\arraystretch}{1.6}
        \centering
        \begin{tabular}{|M{1.6cm}|M{1.7cm}|M{1.7cm}|M{1.7cm}|M{1.7cm}|M{1.7cm}|M{1.7cm}|M{1.7cm}|}\hline
            & \multicolumn{7}{c|}{Fitting error (\%) : Standard NMF / GSVD-NMF} \\\hline
             &  \multicolumn{2}{c|}{LCMS1} & \multicolumn{2}{c|}{LCMS2} & MHLL & \multicolumn{2}{c|}{P\&F No.1} \\\hline 
             $r$&  \multicolumn{2}{c|}{17} & \multicolumn{2}{c|}{23} & 3 & \multicolumn{2}{c|}{12} \\\hline 
             $k$ & 1 & 3& 1 & 4& 1 & 1 & 2 \\ \hline \hline
             \multicolumn{8}{|c|}{HALS} \\\hline
             Random & 3.97$\pm$0.01 / 3.97$\pm$0.00 & 3.97$\pm$0.01 / 3.97$\pm$0.01 & \textbf{4.06$\pm$0.08 / 4.01$\pm$0.06} & \textbf{4.06$\pm$0.08 / 3.99$\pm$0.04} & 3.08$\pm$0.00 / 3.08$\pm$0.00 & \textbf{2.90$\pm$0.11 / 2.85$\pm$0.00} & \textbf{2.90$\pm$0.11 / 2.85$\pm$0.00} \\ \hline
             
             NNDSVD & 3.97 / 3.97 & 3.97 / 3.97 & \textbf{4.05 / 3.99} & \textbf{4.05 / 4.01} & 3.08 / 3.08 & 2.85 / 2.85 & 2.85 / 2.85 \\ \hline
             
             NNDSVDa & 3.97 / 3.97 & 3.97 / 3.97 & \textbf{4.07 / 4.01} & \textbf{4.07 / 4.01} & 3.08 / 3.08 & 2.85 / 2.85 & 2.85 / 2.85 \\ \hline
             
             NNDSVDar & 3.97 / 3.97 & 3.97 / 3.97 & \textbf{4.05 / 3.99} & \textbf{4.05 / 4.01} & 3.08 / 3.08 & 2.85 / 2.85 & 2.85 / 2.85  \\  \hline \hline
             
             \multicolumn{8}{|c|}{GCD} \\\hline
             Random & 3.97$\pm$0.01 / 3.97$\pm$0.00 & 3.97$\pm$0.01 / 3.97$\pm$0.01 & \textbf{4.03$\pm$0.07 / 3.99$\pm$0.06} & \textbf{4.03$\pm$0.07 / 3.98$\pm$0.04} & 3.08$\pm$0.00 / 3.08$\pm$0.00 & \textbf{2.90$\pm$0.11 / 2.85$\pm$0.00} & \textbf{2.90$\pm$0.11 / 2.85$\pm$0.00} \\ \hline
             
             NNDSVD & 3.98 / 3.97 & 3.98 / 3.97 & \textbf{3.99 / 3.94} & 3.99 / 4.01 & 3.08 / 3.08 & 2.85 / 2.85 & 2.85 / 2.85 \\ \hline
             
             NNDSVDa & 3.97 / 3.97 & 3.97 / 3.97 & 4.00 / 4.10 & 4.00 / 4.00 & 3.08 / 3.08 & 2.85 / 2.85 & 2.85 / 2.85 \\ \hline
             
             NNDSVDar & 3.97 / 3.97 & 3.97 / 3.97 & \textbf{3.99 / 3.94} & 3.99 / 4.01 & 3.08 / 3.08 & 2.85 / 2.85 & 2.85 / 2.85  \\  \hline \hline
             
             \multicolumn{8}{|c|}{ALSPGrad} \\\hline
             Random & 3.97$\pm$0.01 / 3.96$\pm$0.00 & 3.97$\pm$0.01 / 3.97$\pm$0.00 & \textbf{4.06$\pm$0.06 / 4.02$\pm$0.04} & \textbf{4.06$\pm$0.06 / 4.02$\pm$0.05} & 3.08$\pm$0.00 / 3.08$\pm$0.00 & \textbf{2.89$\pm$0.10 / 2.85$\pm$0.00} & \textbf{2.89$\pm$0.10 / 2.85$\pm$0.00} \\ \hline
             
             NNDSVD & 3.97 / 3.97 & 3.97 / 3.97 & \textbf{3.99 / 3.94} & 3.99 / 4.01 & 3.08 / 3.08 & 2.85 / 2.85 & 2.85 / 2.85 \\ \hline
             
             NNDSVDa & 3.96 / 3.97 & 3.96 / 3.97 & \textbf{4.19 / 4.01} & \textbf{4.19 / 3.94} & 3.08 / 3.08 & 2.85 / 2.85 & 2.85 / 2.85 \\ \hline
             
             NNDSVDar & 3.97 / 3.97 & 3.97 / 3.97 & \textbf{3.99 / 3.94} & 3.99 / 4.01 & 3.08 / 3.08 & 2.85 / 2.85 & 2.85 / 2.85  \\  \hline \hline
             
             \multicolumn{8}{|c|}{MU} \\\hline
             Random & 4.00$\pm$0.02 / 4.03$\pm$0.01 & 4.00$\pm$0.02 / 4.16$\pm$0.05 & \textbf{4.18$\pm$0.08 / 4.16$\pm$0.06} & 4.18$\pm$0.08 / 4.27$\pm$0.08 & 3.12$\pm$0.06 / 3.39$\pm$0.02 & 2.91$\pm$0.11 / 2.91$\pm$0.04 & 2.91$\pm$0.11 / 3.08$\pm$0.24 \\ \hline
             
             NNDSVD & 5.09 / 5.21 & 5.09 / 5.12 & \textbf{5.89 / 5.82} & \textbf{5.89 / 5.62} & 3.24 / 3.68 & \textbf{4.50 / 4.11} & \textbf{4.50 / 4.15} \\ \hline
             
             NNDSVDa & 4.01 / 4.02 & 4.01 / 4.18 & 4.43 / 4.52 & \textbf{4.43 / 4.30} & 3.09 / 3.47 & 2.85 / 2.91 & 2.85 / 3.64 \\ \hline
             
             NNDSVDar & 4.02 / 4.07 & 4.02 / 4.17 & \textbf{4.14 / 4.03} & 4.08 / 4.30 & 3.09 / 3.45 & 2.86 / 2.91 & 2.86 / 3.64  \\  \hline 
        \end{tabular}
        \caption{GSVD-NMF vs standard NMF with different initialization}
        \label{diff_ini_compare}
    \end{table*}

    \begin{table*}[!t]
    \renewcommand{\arraystretch}{1.6}
        \centering
        \begin{tabular}{|M{2cm}|M{1.7cm}|M{1.7cm}|M{1.7cm}|M{1.7cm}|M{1.7cm}|M{1.7cm}|M{1.7cm}|}\hline
            & \multicolumn{7}{c|}{Time($s$)} \\\hline
             &  \multicolumn{2}{c|}{LCMS1} & \multicolumn{2}{c|}{LCMS2} & MHLL & \multicolumn{2}{c|}{P\&F No.1} \\\hline 
             $r$&  \multicolumn{2}{c|}{17} & \multicolumn{2}{c|}{23} & 3 & \multicolumn{2}{c|}{12} \\\hline 
             $k$ & 1 & 3& 1 & 4& 1 & 1 & 2 \\ \hline \hline
             
             Feature Recovery & \textbf{0.0015$ \pm$ 0.0001} & \textbf{0.0013 $\pm$ 0.0001} & \textbf{0.0020 $\pm$ 0.0001} & \textbf{0.0018 $\pm$ 0.0002} & \textbf{0.0001 $\pm$ 0.0000} & \textbf{0.0011 $\pm$ 0.0001} & \textbf{0.0015 $\pm$ 0.0001} \\ \hline
             
             Final NMF & 0.8561 $\pm$ 0.4939 & 0.7209 $\pm$ 0.2950 & 0.2220 $\pm$ 0.1924 & 0.4750 $\pm$ 0.3691 & 0.0052 $\pm$ 0.0003 & 0.2461 $\pm$ 0.1133 & 0.3125 $\pm$ 0.0229 \\ \hline
             
             Standard NMF & 0.9039 $\pm$ 0.3380 & 0.9039 $\pm$ 0.3380 & 0.8355 $\pm$ 0.3179 & 0.8355 $\pm$ 0.3179 & 0.0063 $\pm$ 0.0024 & 0.3616 $\pm$ 0.1014 & 0.3616 $\pm$ 0.1014  \\  \hline
        \end{tabular}
        \caption{Time (mean $\pm$ std) of each stage of GSVD-NMF}
        \label{time_of_each_part}
    \end{table*}

\end{document}